\documentclass{article}

\usepackage{arxiv}

\usepackage[utf8]{inputenc}
\usepackage[T1]{fontenc}
\usepackage{hyperref}
\usepackage{url}
\usepackage{booktabs}
\usepackage{amsmath,amssymb,amsfonts}
\usepackage{graphicx}
\usepackage{float}
\usepackage{multirow}
\usepackage{nicefrac}
\usepackage{microtype}
\usepackage[numbers]{natbib}
\usepackage{doi}
\usepackage{xcolor}
\usepackage{algorithmicx}
\usepackage{algpseudocode}
\usepackage{listings}
\usepackage{tabularx}
\usepackage{multicol}
\usepackage{cleveref}
\usepackage{siunitx}
\usepackage{subfig}
\usepackage{amsthm}
\usepackage{mathrsfs}
\usepackage[title]{appendix}
\usepackage[ruled,vlined,linesnumbered]{algorithm2e}

\title{Extending Decision Predicate Graphs for Comprehensive Explanation of Isolation Forest}

\author{
  Matteo Ceschin \\
  Department of Engineering and Architecture \\
  University of Trieste, Italy
  \And
  \href{https://orcid.org/0009-0006-2494-0349}{\includegraphics[scale=0.06]{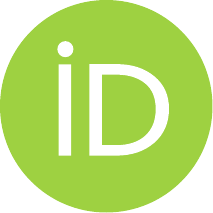}\hspace{1mm}Leonardo Arrighi}\thanks{Corresponding author: \texttt{leonardo.arrighi@phd.units.it}} \\
  Department of Mathematics, Informatics, and Geosciences \\
  University of Trieste, Italy
  \And
  \href{https://orcid.org/0000-0002-2718-5426}{\includegraphics[scale=0.06]{orcid.pdf}\hspace{1mm}Luca Longo} \\
  Centre of explainable artificial intelligence \\
  Artificial intelligence and cognitive load research lab \\
  Technological University Dublin, Ireland
  \And
  \href{https://orcid.org/0000-0002-4988-0702}{\includegraphics[scale=0.06]{orcid.pdf}\hspace{1mm}Sylvio Barbon Junior} \\
  Department of Engineering and Architecture \\
  University of Trieste, Italy
}

\hypersetup{
  pdftitle={Extending Decision Predicate Graphs for Comprehensive Explanation of Isolation Forest},
  pdfsubject={cs.AI, cs.LG},
  pdfauthor={Matteo Ceschin, Leonardo Arrighi, Luca Longo, Sylvio Barbon Junior},
  pdfkeywords={Ensemble Learning, Outliers, Explainable Artificial Intelligence, Interpretability, Anomalies, Tree-based Ensemble Model},
}

\begin{document}
\maketitle

\begin{abstract}
The need to explain predictive models is well-established in modern machine learning.
However, beyond model interpretability, understanding pre-processing methods is equally essential.
Understanding how data modifications impact model performance improvements and potential biases and promoting a reliable pipeline is mandatory for developing robust machine learning solutions. 
Isolation Forest (iForest) is a widely used technique for outlier detection that performs well. Its effectiveness increases with the number of tree-based learners. However, this also complicates the explanation of outlier selection and the decision boundaries for inliers.
This research introduces a novel Explainable AI (XAI) method, tackling the problem of global explainability. In detail, it aims to offer a global explanation for outlier detection to address its opaque nature.
Our approach is based on the Decision Predicate Graph (DPG), which clarifies the logic of ensemble methods and provides both insights and a graph-based metric to explain how samples are identified as outliers using the proposed Inlier-Outlier Propagation Score (IOP-Score).
Our proposal enhances iForest's explainability and provides a comprehensive view of the decision-making process, detailing which features contribute to outlier identification and how the model utilizes them. 
This method advances the state-of-the-art by providing insights into decision boundaries and a comprehensive view of holistic feature usage in outlier identification.---thus promoting a fully explainable machine learning pipeline.
\end{abstract}


\maketitle

\section{Introduction}\label{Intro}

Most current Explainable AI (XAI) techniques predominantly focus on elucidating predictive models, often overlooking the necessity of addressing the entire data processing pipeline. This partial focus can result in incomplete explanations regarding the context, potentially leaving critical aspects of data handling and pre-processing---such as feature selection and outlier removal---obscured. As Lipton \cite{lipton2018mythos} argues, a holistic approach to explainability is essential for the credibility and utility of machine learning solutions. Similarly, authors advocate for a shift towards transparent machine learning ecosystems, where every pipeline component, from data preprocessing to model decision-making, is made transparent \cite{holzinger2020measuring,ViloneLongo2023}. More robust, trustworthy explanations can be constructed by ensuring XAI techniques encompass the entire pipeline.
Data preparation and transformation models before training a predictive model demand clarity equal to the last one for several reasons, including transparency, reliability, and regulatory requirements \cite{sipple2022general}.
Firstly, transparency in pre-processing enhances the understandability of the data manipulations that occur before model training \cite{strasser2024transparent}. 
By understanding how data is cleaned, normalized, and selected during pre-processing, users can identify potential sources of bias or errors that might affect the model’s performance. Furthermore, this process enables the detection and mitigation of data acquisition issues, such as systematic errors or noise, and supports enhancements to the overall system pipeline. Finally, clear documentation and explanation of all stages of data handling, including pre-processing, ensure compliance with these regulations and promote trust and reliability \cite{zelaya2019towards}.\\

Among many pre-processing algorithms, Isolation Forest (iForest) \cite{liu2008isolation} stands out due to its straightforward approach and effectiveness in swiftly handling outliers in high-dimensional data. However, the core mechanism of iForest, which relies on a random selection of features and split points to isolate anomalies, introduces stochasticity that can sometimes lead to ambiguous or non-intuitive results \cite{Kartha2021}. Consequently, providing explanations for the decisions made by iForest is essential, as it allows users to understand and trust the logic behind the identification of outliers, mainly when dealing with complex datasets. These explanations not only help validate the anomalies detected by iForest but also aid in fine-tuning the model by revealing potential biases or errors introduced by the randomness in the selection process \cite{Ndao2024}. 
Shapley Additive exPlanations (SHAP) \cite{Lundberg2017} is currently used to explain the behaviour of the iForest model by providing insights into how features influence its predictions. 
In contrast, the Depth-based Isolation Forest Feature Importance (DIFFI) \cite{carletti2023interpretable} method employs a tailored approach that leverages the internal structure of iForest to compute feature importances.
However, both methods provide a local explanation that uses a feature importance vector to illustrate the model's decision-making process for identifying individual samples. 
While effective, these approaches mainly focus on feature-level contributions without exploring the structural or logical complexities of the iForest ensemble.

To overcome the limitation of providing only a vector of feature importances, we propose a method based on Decision Predicate Graphs (DPG) \cite{arrighi2024decision} to elucidate the logic and intrinsic aspects of the iForest ensemble. 
Building on the principles of the DPG technique, our method converts the iForest model into a graph, allowing us to exploit its structural properties and leverage established mathematical theories to elucidate the outlier detection process.
According to Speith \cite{speith_review_2022}, the proposed method is global, as it explains the entire decision-making process of the iForest model, revealing general patterns and the feature interactions that drive the whole model's logic.
This approach provides a mixed-type explanation, as done in other research \cite{RizzoL18Explainability,makridis2024enhancing}, by integrating a visual representation of the model’s decision-making process with a new quantitative metric, the Inlier-Outlier Propagation Score (IOP-Score),  which assesses each feature’s contribution to outlier detection. 
By extracting relationships and decision paths within the ensemble, our method enhances model transparency and delivers actionable insights into its internal mechanisms, surpassing traditional explanation techniques. Our work contributes in the following ways:
\begin{itemize} 
 \item Comprehensive global explanation of iForest: we propose a method to explain the iForest model, including details on feature boundaries for both inliers and outliers samples. 
\item  The IOP-Score: a novel metric that quantifies a node's tendency to propagate toward either the inliers or outliers to enhance interpretability by distinguishing discriminative from neutral predicates in the iForest.
\item Graph-based interpretability: by integrating DPG, we introduce a graph-based structure that models the isolation logic, such as feature influence on isolation depth and decision paths, enabling a detailed understanding of the detection process. 
\end{itemize}

The results are derived from synthetic and well-established datasets to demonstrate the method's potential. However, we emphasize that the approach is generalizable, indicating its broad applicability across various related scenarios.\\

The remainder of this manuscript is structured as follows: Section 1 introduces the need for transparency in predictive models and pre-processing methods such as outlier detection with iForest. Section 2 provides a background on iForest, existing explanation techniques, and the DPG approach. Section 3 presents the DPG-based explanation framework for iForest, detailing the graph construction process and introducing the IOP-Score. Section 4 describes the experimental setup. Section 5 discusses the approach's limitations and potential extensions, addressing scalability and further improvements. Finally, Section 6 concludes the study, summarizing key contributions and outlining future research directions.

\section{Background and Related Work}\label{LR}

Our background and related work section presents the foundations of the iForest algorithm, the current research on explaining iForest, and finally, a subsection about how DPG works and why we proposed a solution based on this approach.

\subsection{Isolation Forest}
One of the most widely used algorithms for anomaly detection is Isolation Forest, also known as iForest, a tree-based method introduced by Liu \cite{liu2008isolation}. 
iForest is designed to efficiently identify outliers, data points that deviate significantly from other instances of the dataset, instead of inliers, representing most of the data and conforming to expected patterns.
Among the various techniques available, iForest stands out for its efficiency and scalability, thanks to its linear time complexity and low memory consumption. Another key advantage is that iForest is an unsupervised learning method that does not require labelled data for training. Moreover, through an effective subsampling procedure, iForest mitigates the swamp effect, where regular points are wrongly identified as anomalies, and addresses the masking issue, which occurs when multiple anomalies conceal each other.
iForest identifies outliers by recursively partitioning the data. Its core idea is that anomalies are rare and distinct from normal instances, requiring fewer random splits to isolate in the problem space. 
This characteristic enables the algorithm to separate anomalous data points from the majority of inliers efficiently.\\

Given a dataset \(X\), where \(d\) features characterize each instance, the iForest consists of multiple binary trees, called Isolation Trees (iTrees), that form the forest.
Each tree is built by randomly selecting a feature \(d_i\) and a random value \(v\) within the range \([ \min(v_{d_i}), \max(v_{d_i}) ]\), where $v_{d_i}$ are the values of the samples of $X$ associated to the feature $d_i$. 
If an instance's selected feature value \(v_{d_i}\) is less than \(v\), the instance is directed to the left branch; otherwise, it is directed to the right branch of the iTree.
After each split, the dataset is partitioned so each branch contains a subset of $X$.
This process is recursively applied to the resulting subsets until one of the following stopping conditions is met:
\begin{itemize}
    \item The iTree reaches its maximum depth, which is defined as:
    \[
    \lceil \log_2(\min(256, |X|)) \rceil,
    \]
    where $|X|$ is the number of samples of the dataset.
    This ensures that the tree does not grow indefinitely. 
    \item A single instance has been completely isolated in a leaf node.
    \item Two or more identical instances have been grouped into a single leaf node, making further splits impossible.
\end{itemize}

Once an iTree is fully grown, each instance $x$ in \(X\) is assigned to a leaf node. Its path length \(h(x)\) is the number of edges traversed from the root to that leaf.
This recursive process is repeated \(n\) times to build \(n\) trees in the forest.
The final step of the iForest algorithm is the calculation of the \emph{anomaly score} for each instance in the dataset. This score allows the model to determine whether a sample is an outlier (anomaly) or an inlier.
The anomaly score is computed as follows:
\[
s(x, n) = 2^{-\frac{\mathbb{E}(h(x))}{c(n)}},
\]
where \(\mathbb{E}(h(x))\) is the average path length of \(x\) across all trees in the forest, and \(c(n)\) is a normalization factor that estimates the average path length required to isolate a data point in a binary search tree containing 
n instances and is given by:
    \[
    c(n) = 2H(n - 1) - \frac{2(n-1)}{n}
    \]
    where \(H(i)\) is the harmonic number, and it can be estimated by \(ln(i) + \ 0.5772\) (Euler's constant).
If \(s(x, n) < 0.5\), then \(x\) is likely to be a typical instance (inlier). Conversely, if \(s(x, n)\) is close to 1, then \(x\) is highly likely to be an outlier.
The core idea behind the iForest algorithm is that outliers require fewer partitions to be isolated, resulting in shorter path lengths than inliers.

\subsection{Explaining Isolation Forest model}
The literature presents several post-hoc XAI methods designed to interpret the iForest model.
Post-hoc XAI methods are applied after training to provide interpretability without altering the model's internal structure, thereby preserving its performance.
According to Speith \cite{speith_review_2022}, we can distinguish the model-agnostic XAI methods, such as SHapley Additive exPlanations (SHAP) \cite{Lundberg2017}, which can be applied independently of the underlying model, from the model-specific method, tailored for specific models or model classes.
Considering proposals using SHAP, Rachwał et al. \cite{Rachwal2023} proposed an improved iForest algorithm that dynamically excludes attributes based on SHAP indices, resulting in enhanced prediction accuracy and better feature selection. In their approach, SHAP values are used to quantify the importance of each feature, and models are iteratively trained with one feature excluded at a time. The final anomaly score of iForest is computed as a weighted average of these models' anomaly scores, where the weights are derived from the absolute SHAP values, prioritizing features with higher SHAP values and reducing the influence of less relevant ones.
Liu and Aldrich \cite{Liu2023} introduced the iForest-RF-SHAP framework, a novel approach for anomaly detection and explanation in coal data, which combines iForest, Random Forest, and SHAP. This framework outperformed traditional methods, such as principal component analysis, while offering detailed insights into variable contributions. 
In contrast, the main model-specific proposals specifically tailored to explain iForest models include the methods introduced in \cite{Kartha2021,carletti2023interpretable,arcudi2024enhancing}. 
Kartha et al. \cite{Kartha2021} developed a method specifically designed to explain iForest anomaly predictions by assigning a vector of feature importance weights to each attribute, indicating its contribution to the anomaly detection process. These weights are computed by analyzing how much each attribute contributes to isolating a data point within the iForest trees, with higher weights associated with shorter path lengths. The result is an explanation vector that reflects the relative importance of each feature in determining the anomaly score. 
Arcudi et al. \cite{arcudi2024enhancing} introduced Extended Isolation Forest Feature Importance (ExIFFI), a method designed to deliver global and local explanations for iForest. ExIFFI uses feature importance metrics to explain anomaly detection comprehensively, offering a detailed perspective on how individual features contribute to the model's predictions. The feature importance metrics are computed by analyzing the projections of the hyperplane’s normal vector at each node in the isolation trees and weighting them based on the degree of imbalance in the data split, favoring nodes where the sample falls into the smaller partition, thus attributing greater importance to features that isolate anomalies more effectively.
Carletti et al. \cite{carletti2023interpretable} presented Depth-based Isolation Forest Feature Importance (DIFFI), a method tailored for iForest. DIFFI provides global and local interpretability by analyzing how features influence the depth at which anomalies are isolated in the decision trees. This method explains the anomaly detection process and enables unsupervised feature selection, a valuable tool for handling high-dimensional data in anomaly detection problems. 
Despite the advancements in explaining iForest models using methods like SHAP, ExIFFI, and DIFFI, a significant gap remains in providing detailed interpretability regarding the values, intervals, and specific characteristics of inliers alongside outliers. This lack of explanation motivates the development of the DPG-based method, which aims to address these limitations.

\subsection{Decision Predicate Graph}\label{sec:DPG}

\begin{figure}[h!]
    \centering
    \includegraphics[width=\linewidth]{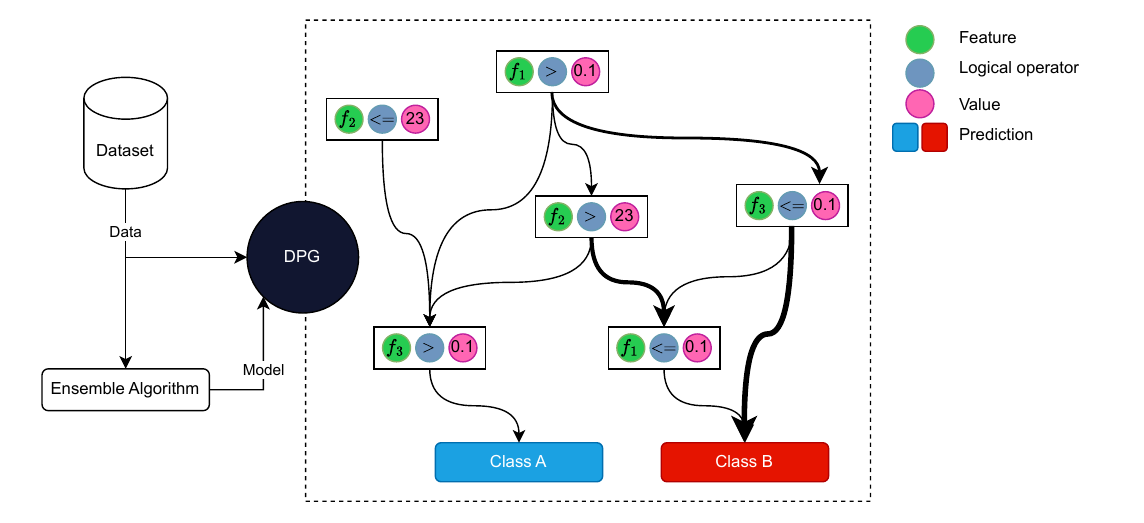}
    \caption{Schematic demonstration of how DPG works.}
    \label{fig:howpdgoworks}
\end{figure}

Decision Predicate Graph (DPG) is a post-hoc, model-specific XAI technique for interpreting tree-based ensemble models.
The DPG method transforms the ensemble model into a weighted directed graph representing the entire decision-making process underlying the model.
It then introduces graph-theoretic metrics that highlight key features of the ensemble model.
After training the tree-based ensemble model, the internal nodes of each base learner, which contain the dataset's split rules, are used to construct predicates---feature-value associations expressed as logical statements (e.g., ``$f_i$ $>$ $v$'', where $f_i$ is a generic feature and $v$ is the associated value in the split). 
The predicates are the graph nodes, as Figure \ref{fig:howpdgoworks} depicts.

Then, each training sample traverses the decision trees again. 
A node is connected to another if, during traversal, a data sample first satisfies the predicate of the initial node and then subsequently satisfies the predicate of the next node.
The graph's edges represent the frequency predicates consecutively satisfied by the training samples during the model’s training phase.
The result is a global explanation of the model, comprising two main components: a visual representation of the decision-making process as a graph illustrating the entire structure.

The visual representation of iForest through DPG highlights how outliers and inliers are treated within the model. Unlike traditional feature importance methods, which provide a vector-based ranking, DPG could leverage graph structures to uncover decision paths, feature interactions, and hierarchical dependencies. This approach enhances interpretability by identifying which features contribute to an anomaly and explaining how and why those features lead to an outlier classification.

\section{DPG-based explanation for Isolation Forest}\label{sec:pseudo_code}

\begin{figure}[h!]
    \centering
    \includegraphics[width=1\linewidth]{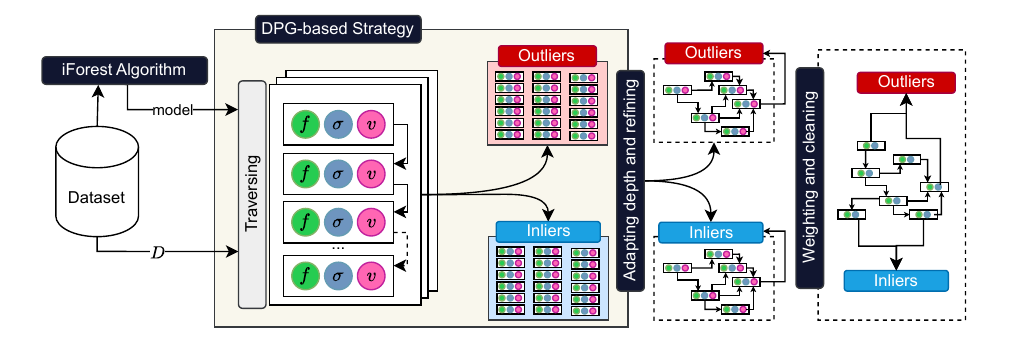}
    \caption{Overview of the proposed approach: iForest DPG representation. Predicates are represented as triples ($f$, $\sigma$, $v$) and are color-coded (green, blue, and pink).}
    \label{fig:overview}
\end{figure}

We propose a novel post-hoc method based on DPG, a model-specific XAI technique designed to understand the decision-making process of the iForest model. An overview of our proposed approach can be seen in Figure \ref{fig:overview}. Subsection \ref{DPG_inter} provides an in-depth explanation of each step.

This technique builds upon the construction method of the DPG, transforming the iForest into a graph structure.
The method captures the inner logic of the iForest model, emphasizing the key decisions and the most frequently used features for identifying outliers. 
It provides a comprehensive global mixed-type explanation by combining a visual representation of the model’s entire decision-making process, depicted as a graph, with a metric that quantifies the importance of each feature in detecting outliers. 
In this section, we detail the construction of our technique and present an in-depth explanation of its components. Additionally, we discuss the necessity of this technique, its advantages, and the key insights it offers into the model’s behavior.

\subsection{Proposed Global Explainability}\label{DPG_inter}

\paragraph{Applying iForest:} To construct the explanation, we begin with the iForest model trained on the dataset. The objective is to comprehend the model’s decision-making process and identify features differentiating inliers from outliers.
The model's output consists of the observations classified as outliers.
These observations are assigned labels: ``Outlier'' if the model classifies them as such, and ``Inlier'' otherwise.

\paragraph{DPG-based strategy:}
Following the DPG proposal, we examine the internal nodes of each tree-based learner in iForest, which contain the dataset's split rules used to construct the predicates defined in DPG.
These predicates are represented as triples ($f$, $\sigma$, $v$), where the sign ($\sigma$) can be either $>$ or $\leq$.
Subsequently, each training sample traverses each tree.
We identify all predicate lists satisfied by the samples in each tree-based learner.
Each list is then extended by appending the label previously assigned to the observations: ``Outlier'' if the list results from an outlier's traversal of the tree, and ``Inlier'' otherwise.
As a result, each observation is associated with a set of predicate lists.

\paragraph{Adapting to an iForest DPG:}
To align with the principles of iForest, which classifies observations that reach the maximum tree depth as inliers, we eliminate all predicate lists that exceed the trees' maximum depth from the outlier sets.
This step is crucial because iForest identifies outliers based on their early isolation, i.e., when an observation becomes separated in a leaf before reaching the maximum depth.
Since observations that reach this depth may not be truly isolated or may not exhibit outlier characteristics, their removal prevents ambiguity that could lead to their misclassification as inliers.

After generating the predicate lists, we further refine them by removing the values ($v$) from each predicate triple, resulting in pairs of the form ($f$, $\sigma$). From now on, we will refer to these pairs as predicates.

This abstraction is necessary because iForest selects the split value ($v$) randomly at each node and for each tree. As a result, the exact triples ($f$, $\sigma$, $v$) are typically unique to individual trees and are not shared or reused across trees. Aggregating predicates at the level of ($f$, $\sigma$, $v$) would therefore hinder cross-tree analysis and reduce the generalizability of the method. By focusing on the feature and direction of the split only, we retain a meaningful and aggregable representation of the isolation patterns across trees.

\paragraph{Weighting iForest DPG:}
Using the predicate lists, we construct a weighted directed graph that represents the entire model.
The predicates serve as the nodes of the graph. 
A node is connected to another if, within the predicate lists, the predicate in the first node is immediately followed by the predicate in the second node. 
This ensures that the connection represents the sequential order in which the predicates are satisfied during a decision tree's traversal.
The graph's edges represent the frequency with which the pair of predicates stored in the connected nodes appears consecutively in the predicate lists, with the order preserved.
The resulting graph shows two classes: ``Outlier'' and ``Inlier'', with their respective predicates distinguished by their frequency and position within the model logic.

\paragraph{Cleaning iForest DPG:}
We can observe that when there is a significant imbalance between the number of outliers and inliers, adjusting the frequency calculation becomes necessary to ensure a fair comparison between the two classes.
Predicates satisfied by outliers appear considerably less frequently than those satisfied by inliers.
Consequently, identifying the distinctive predicates of each class becomes particularly challenging due to the low frequency of those associated with outliers.
We, therefore, introduce a weighting system for the frequencies. 
For each dataset instance that traverses the model, the transition between two consecutive predicates contributes differently depending on the class assigned to the data point.
If the instance is classified as an outlier, its contribution to the frequency is multiplied by a weight $w_o$. Otherwise, its contribution is multiplied by a weight $w_i$. The weights are defined as:
\begin{equation}
\begin{gathered}
    w_o = \frac{N_o + N_i}{N_o}, \quad
    w_i = \frac{N_o + N_i}{N_i},    
\end{gathered}
\end{equation}
where $N_o$ and $N_i$ denote the number of outliers and inliers in the dataset, respectively.
We can, therefore, state that the transition between two consecutive predicates satisfied by an outlier has a weighted frequency equal to $w_o$, while that satisfied by an inlier has a weighted frequency equal to $w_i$. 
The weight of an edge is calculated as the sum of these weighted frequencies; for brevity, we refer to this sum as the \emph{weighted frequency of the edge}.
\paragraph{Towards Explanation.}
Once the graph is constructed, we define a new metric called the \emph{Inlier-Outlier Propagation Score} (IOP-Score), quantifying a node's tendency to lead toward either the ``Outlier'' or ``Inlier'' class. 
This score is calculated as the difference between the frequency of data transitions from a node toward the  ``Inlier'' class and those toward the ``Outlier'' class, normalized by the total frequency of data transitions entering the node. This normalization ensures the score accounts for the node's overall context, providing a balanced measure of its tendency to propagate toward either class.
So, the IOP-Score for a generic node $v$ is defined as:
\begin{equation}
    \text{IOP-Score}(v) = \frac{f_i(v) - f_o(v)}{f_{in}(v)},
\end{equation}
where $f_i(v)$ is the frequency of the edge connecting node $v$ to the ``Inlier'' class, $f_o(v)$ is the frequency of the edge connecting node to the ``Outlier'' class, and $f_{in}(v)$ is the sum of the frequencies of all edges entering node $v$.
\begin{itemize}
    \item If $\text{IOP-Score}(v)=1$, the node is fully associated with the ``Inlier'' class, meaning its frequency results exclusively from transitions toward the ``Inlier'' class. In other words, the predicate appears only in predicate lists generated by inliers traversing the model. 
    \item If $\text{IOP-Score}(v)=-1$, the node is entirely associated with the ``Outlier'' class, with its frequency stemming solely from transitions toward the ``Outlier'' class, indicating that the predicate appears only in predicate lists generated by outliers.
    \item If $\text{IOP-Score}(v)=0$, the node is considered neutral, as there is an equal frequency of transitions toward both the ``Inlier'' and ``Outlier'' classes. 
\end{itemize}

In summary, an \text{IOP-Score} close to $0$ indicates that the node is non-discriminative, while values near $1$ or $-1$ signify predicates that strongly characterize one of the two classes. \\

Outlined in Algorithm \ref{algpseudocode}, the proposed approach is presented in pseudocode to enhance clarity and understanding.

\begin{algorithm}[H]
    \SetKwFunction{computeWeights}{ComputeFrequencyWeights}
    \caption{iForest as a graph for DPG-based Explanation}
    \label{algpseudocode}
    \KwIn{Trained Isolation Forest model $IF$, Dataset $D$, maximum depth of trees $dmax$}
    \KwOut{iForest $DPG$}
    
    Initialize empty graph $G$\;
    \ForEach{base learner iTree ($iT$) in $IF$} {
        Extract split rules defining predicates $(f, \sigma, v)$\;
        \ForEach{training sample $s$ traversing $iT$} {
            Record satisfied predicate lists\;
            \If{$s$ classified as outlier} {
                Label list as ``Outlier''\;
            }
            \Else {
                Label list as ``Inlier''\;
            }
        }
    }
    Remove paths of predicates exceeding tree $dmax$ for outliers\;
    Transform predicate lists to pairs $(f, \sigma)$\;
    \ForEach{predicate pair $(p_i, p_j)$ appearing consecutively} {
        Create directed edge $(p_i \rightarrow p_j)$ with frequency weight\;
    }
    Apply class-based frequency weighting using \computeWeights{}\;
    \Return $G$\;
\end{algorithm}

\subsection{Understanding the Explanation Process}
The proposed technique is designed to capture the key concept underlying iForest. 
In iForest, outliers are isolated more rapidly than inliers, requiring fewer splits to separate them from the rest of the dataset.
Although selecting features and associated values at each split is random, outliers differ from inliers for certain features. 
These key features play a role in the splits that lead to the isolation of outliers.
Our XAI method's purpose is to identify the features that differentiate outliers from inliers and understand their role in IF’s decision-making process.
Representing the process as a graph enables visualization of predicate sequences leading to each class, highlighting the typical paths of outliers.
By incorporating information about the sign of the predicates, the method enables the interpretation of the direction of the constraints imposed by the model---that is, whether a feature contributes to the isolation of outliers by surpassing a certain threshold.
Moreover, using the IOP-Score---calculated for each node of the graph---quantifies the relative contribution of features in distinguishing between the two classes.
A low value of this metric indicates that the corresponding predicate is essential for isolating outliers, emphasizing that the outlier nature of observations depends on specific features. 
This aspect underscores the importance of correctly interpreting these features within the context of the application domain and the need to consider potential data errors that may affect the identification of outliers.
Furthermore, the weight of the edges connecting the nodes---proportional to frequencies---indicates whether the predicates are immediately effective at distinguishing outliers, such as when outliers are easily separated along a feature, or whether they contribute indirectly by forming decision paths that require additional splits to isolate an outlier.
By combining the graph structure with IOP-Score, the proposed technique provides a global and interpretable explanation of the model. It illustrates not only which features are used but also how and with what frequency they contribute to isolating outliers.

\Cref{tab:dpg_graph} summarizes how to interpret the DPG structure to understand its implications for outlier and inlier classification.

\begin{table}[ht!]
    \footnotesize
    \centering
    \caption{DPG and their Implications for Outlier/Inlier interpretation}
    \label{tab:dpg_graph}
    \begin{tabular}{p{4cm}p{7cm}}
        \toprule
        \textbf{Component} & \textbf{Implication for Outlier/Inlier Detection} \\
        \midrule
        \textbf{Node (Predicate)} & Represent a decision made to identify a sample as an inlier or outlier pathway based on feature and condition. \\
        \midrule
        \textbf{Weighted Edge} & Indicates how frequently a decision path is used. Thicker edges leading to outliers highlight important anomaly detection features. \\
        \midrule
        \textbf{Node (Terminal)} & Base on classified samples as inliers or outliers, helping identify critical predicates for anomaly separation. \\
              \midrule
        \textbf{IOP-Score} &  Predicates with negative IOP-Scores correspond to features that play a major role in isolating outliers, while positive values indicate features that help define inlier boundaries. \\
       
        \bottomrule
    \end{tabular}
\end{table}

\section{Experiments}

This section demonstrates the novelty and contributions of our DPG-based approach to explaining the iForest. 
We utilized a synthetic dataset to construct challenging anomalies featuring multiple attributes across various scales. 
Additionally, we employed a benchmark dataset to facilitate a fair comparison with other techniques. 
This benchmark dataset was also used in the original iForest study. 
We conducted a comprehensive analysis, utilizing both visualizations and interpretations provided by our method.

Our implementation was developed in Python, leveraging a suite of libraries to facilitate anomaly detection, visualization, and data processing. The scikit-learn library \cite{scikit-learn} was utilized for the implementation of the iForest algorithm, while Graphviz enabled the visualization of the DPG\footnote{Implementation available at: \url{https://github.com/LeonardoArrighi/DPG}}
, enhancing the interpretability of the decision-making process. To promote reproducibility and facilitate further research, the complete source code is publicly available on GitHub\footnote{Implementation available at: \url{https://github.com/Math0097/DPG-iForest}}.

\subsection{Synthetic datasets}

To analyze our XAI methods, we generated two synthetic datasets. Each dataset contains \num{200} data points characterized by six numerical features (denoted as $F_i$, where $i$ ranges from \num{1} to \num{6}), all forming a single-cluster distribution.
We introduced outliers by randomly selecting samples and modifying specific feature values according to predefined rules.
Each outlier is generated by altering two or four feature values from a randomly selected sample among the available ones. 
Each alteration is performed by rescaling the original value by a factor of \num{4} or \num{5} times the standard deviation of that feature computed over the entire dataset.
The resulting dataset exhibits clearly defined anomalies, distinct feature variations, and a balanced level of complexity, making it well-suited for assessing explanation techniques in anomaly detection.
We trained an iForest model with \num{200} trees for each study case to identify outliers. Since our focus is on XAI---where the primary objective is to explain the model's decisions rather than optimize predictive accuracy---the exact number of trees is not relevant to our scope. Therefore, we chose \num{200} trees to ensure robust and stable predictions.

\subsubsection{Synthetic dataset with one outlier.}
\begin{figure}[h!]
    \centering
    \includegraphics[width=1\linewidth]{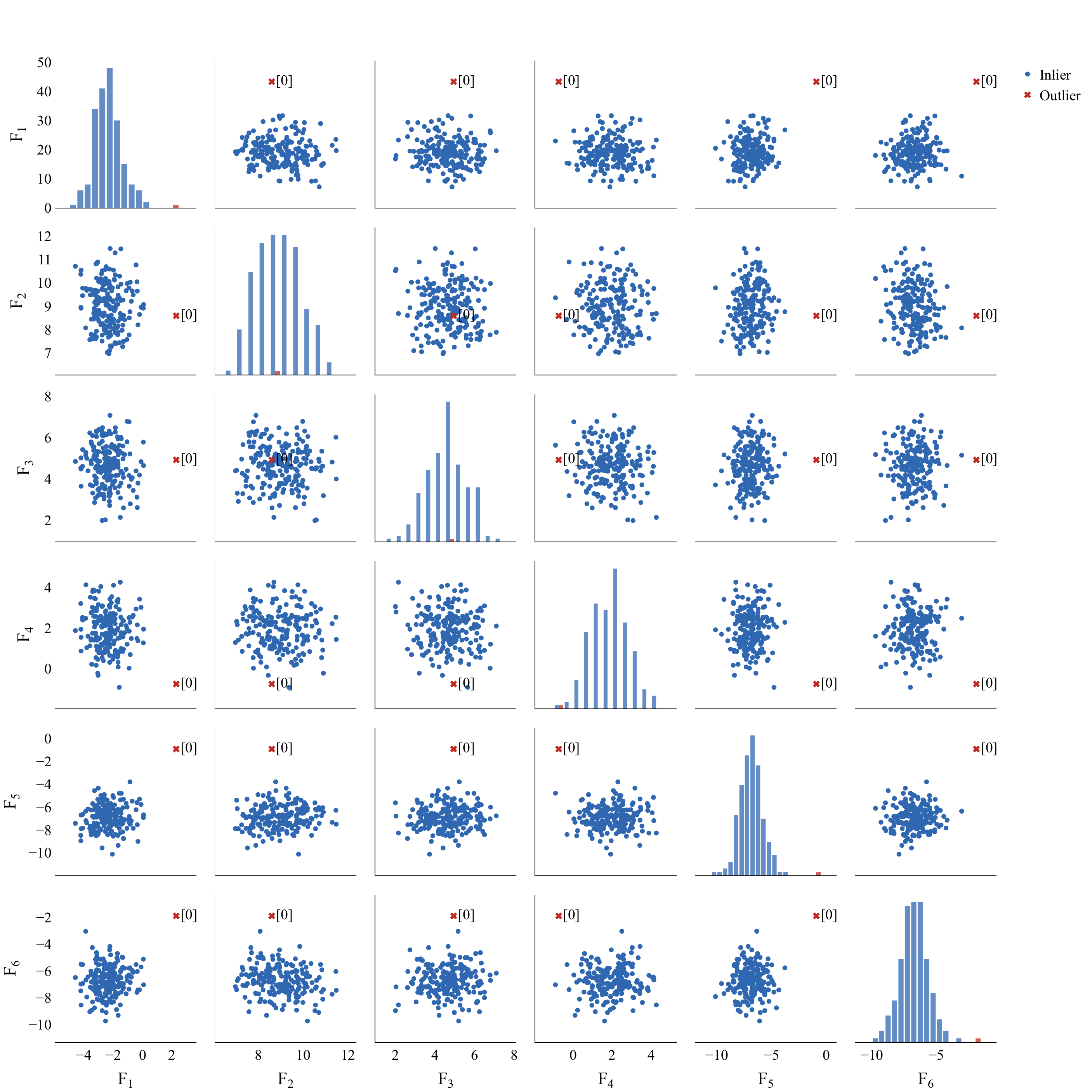}
    \caption{Pairplot of the first synthetic dataset. The dataset comprises \num{200} samples with six numerical features and one outlier.}
    \label{fig:pp_d1_overview}
\end{figure}

The first dataset was generated by modifying four features of one sample, as reported in the \Cref{tab:modifiche_d1}, thereby producing a single outlier among \num{200} samples. 
In \Cref{fig:pp_d1_overview}, we present a pair plot of the first synthetic dataset, where we can observe that the single outline stands apart from the clustered inliers.

\begin{table}[ht!]
\centering
\begin{tabular}{c@{\hspace{0.5cm}}c@{\hspace{0.5cm}}c@{\hspace{0.5cm}}c@{\hspace{0.5cm}}c}
    \toprule
    Outliers & Feature & Initial Value & Final Value & Alteration \\
    \midrule
\multirow{4}{*}{Sample 0} & $F_0$ & $-2.12$ & $2.29$  & $+4.41$ \\
                & $F_3$ & $4.05$ & $-0.76$ & $-4.81$ \\
                & $F_4$ & $-6.01$ & $-0.93$ & $+5.08$ \\
                & $F_5$ & $-7.21$ & $-1.88$ & $+5.33$ \\
    \bottomrule
\end{tabular}
\caption{Sample 0 is the outlier in the first synthetic dataset. The table presents both the initial and final values of the modified features for this sample, along with the specific modifications applied to introduce the outlier.}
\label{tab:modifiche_d1}
\end{table}

\begin{figure}[h!]
    \centering
    \begin{minipage}[t]{0.9\textwidth}
        \includegraphics[width=\textwidth, keepaspectratio]{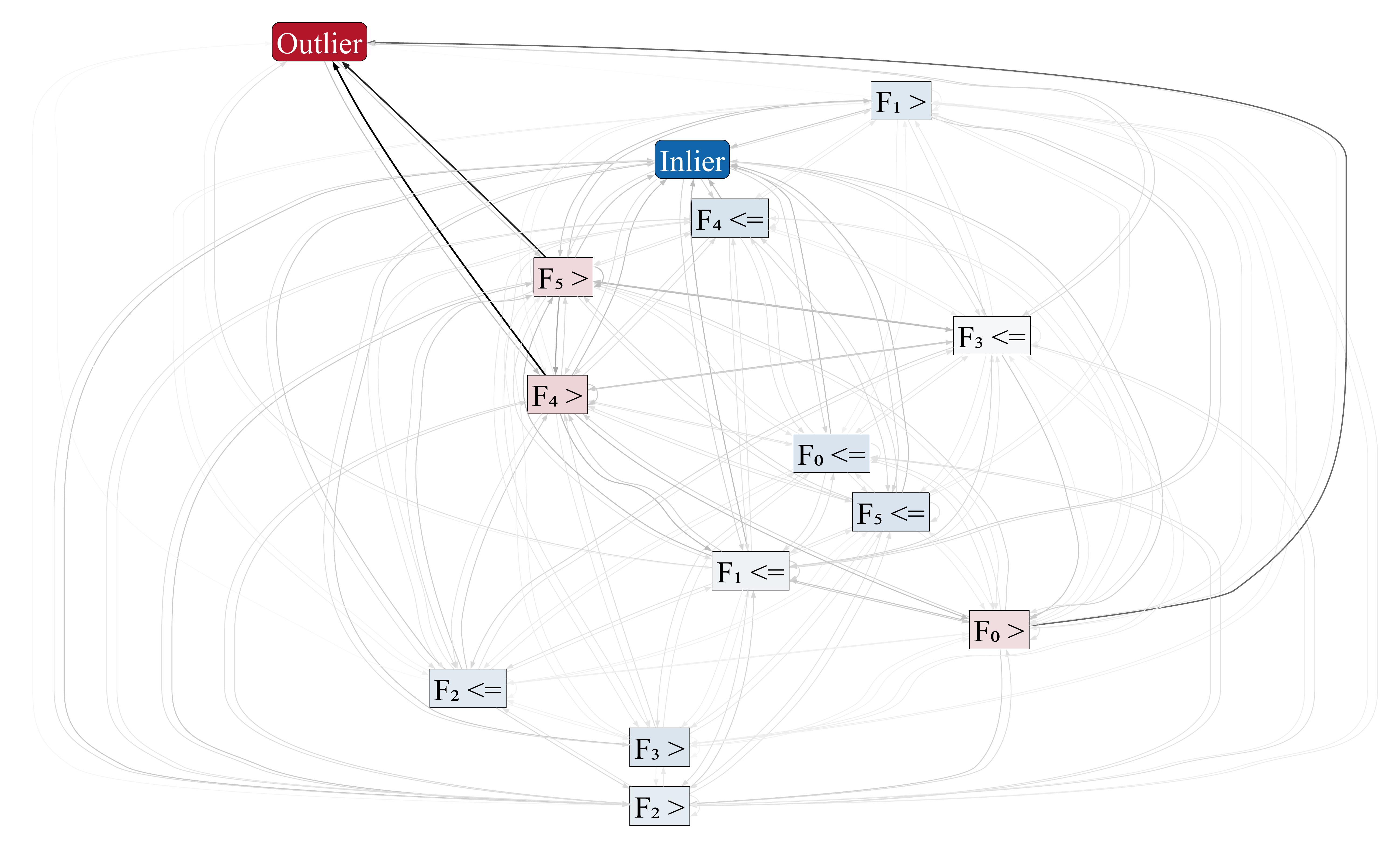}
        \includegraphics[width=\textwidth, keepaspectratio]{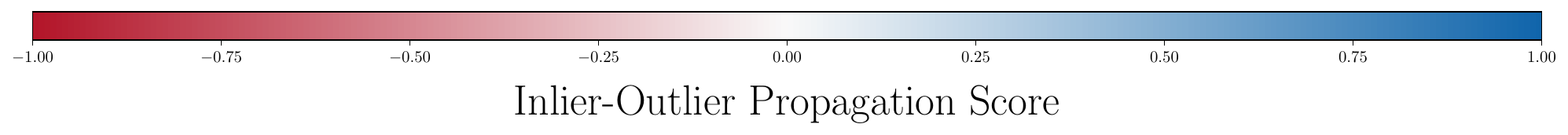}
    \end{minipage}
    \begin{minipage}[t]{0.092\textwidth}
        \raisebox{0cm}{\includegraphics[width=0.9\textwidth, height=8cm, keepaspectratio]{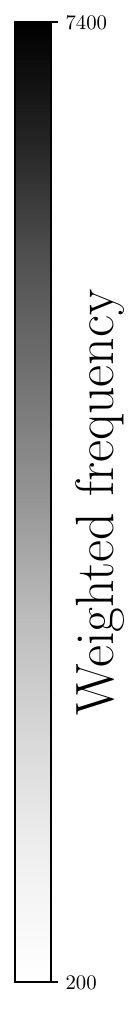}}
    \end{minipage}
    
    \caption{Global representation of the iForest model as a DPG produced by our method for the first synthetic dataset. The vertical bar on the right indicates the edge weights, while the horizontal bar at the bottom displays the IOP-Score of the nodes.}
    \label{fig:global_sintetic_fig_d1}
\end{figure}

The modified sample was correctly identified as an outlier by the iForest model. Then, applying our technique, we obtained iForest DPG, as shown in \Cref{fig:global_sintetic_fig_d1}, where the classes outliers and inliers are distinguished by different colors. For each node, the IOP-Score was computed and represented by its color---these scores are summarized in \Cref{tab:global_sintetic_tab_d1}.

\begin{table}[ht!]
    \centering
    \begin{tabular}{l@{\hspace{0.5cm}}r} 
        \toprule
        Predicate & IOP-Score \\
        \midrule
        $F_4$ $<=$ & $0.1427$ \\
$F_0$ $<=$ & $0.1406$ \\
$F_5$ $<=$ & $0.1336$ \\
$F_3$ $>$ & $0.1304$ \\
$F_1$ $>$ & $0.1129$ \\
$F_2$ $<=$ & $0.0985$ \\
$F_2$ $>$ & $0.0807$ \\
$F_1$ $<=$ & $0.0426$ \\
$F_3$ $<=$ & $0.0091$ \\
$F_0$ $>$ & $-0.1202$ \\
$F_5$ $>$ & $-0.1362$ \\
$F_4$ $>$ & $-0.1580$ \\
        \bottomrule
    \end{tabular}
    \caption{IOP-Score values assigned to each predicate (node) extracted from the DPG graph of the iForest model for the first synthetic dataset. The scores quantify a node’s propensity to distinguish data toward the inliers (positive values) or outliers (negative values) class.}
    \label{tab:global_sintetic_tab_d1}
\end{table}


By examining an in-depth view of the iForest model's internal process, we can observe that some nodes exhibit IOP-Score values below \num{0}, indicating an association with the ``Outliers'' class. The nodes with the lowest scores contain particularly meaningful predicates---namely, $F_4 >$, $F_5 >$, and $F_0 >$---which correspond to the features altered to create the anomaly.
The $>$ sign indicates that, for the outlier, these feature values exceed those of inliers, a fact further supported by the \Cref{fig:pp_d1_overview}. 
Moreover, the edges connecting these nodes to the ``Outlier'' class are thicker, reflecting higher weighted frequencies; this suggests that the model consistently employs splits based on these predicates as final decision points to isolate outliers. 
In contrast, nodes involving predicates on $F_3$, despite it being one of the modified features, do not have low IOP-Score values and are not closely associated with the ``Outlier'' class. 
This indicates that $F_3$ does not consistently separate the anomalous sample from inliers, though it does contribute to the isolation process on several occasions. 
Finally, the remaining nodes with IOP-Score values above \num{0} are predominantly involved in splits that classify points as inliers.

\subsubsection{Synthetic dataset with four outliers.}
\begin{figure}[h!]
    \centering
    \includegraphics[width=1\linewidth]{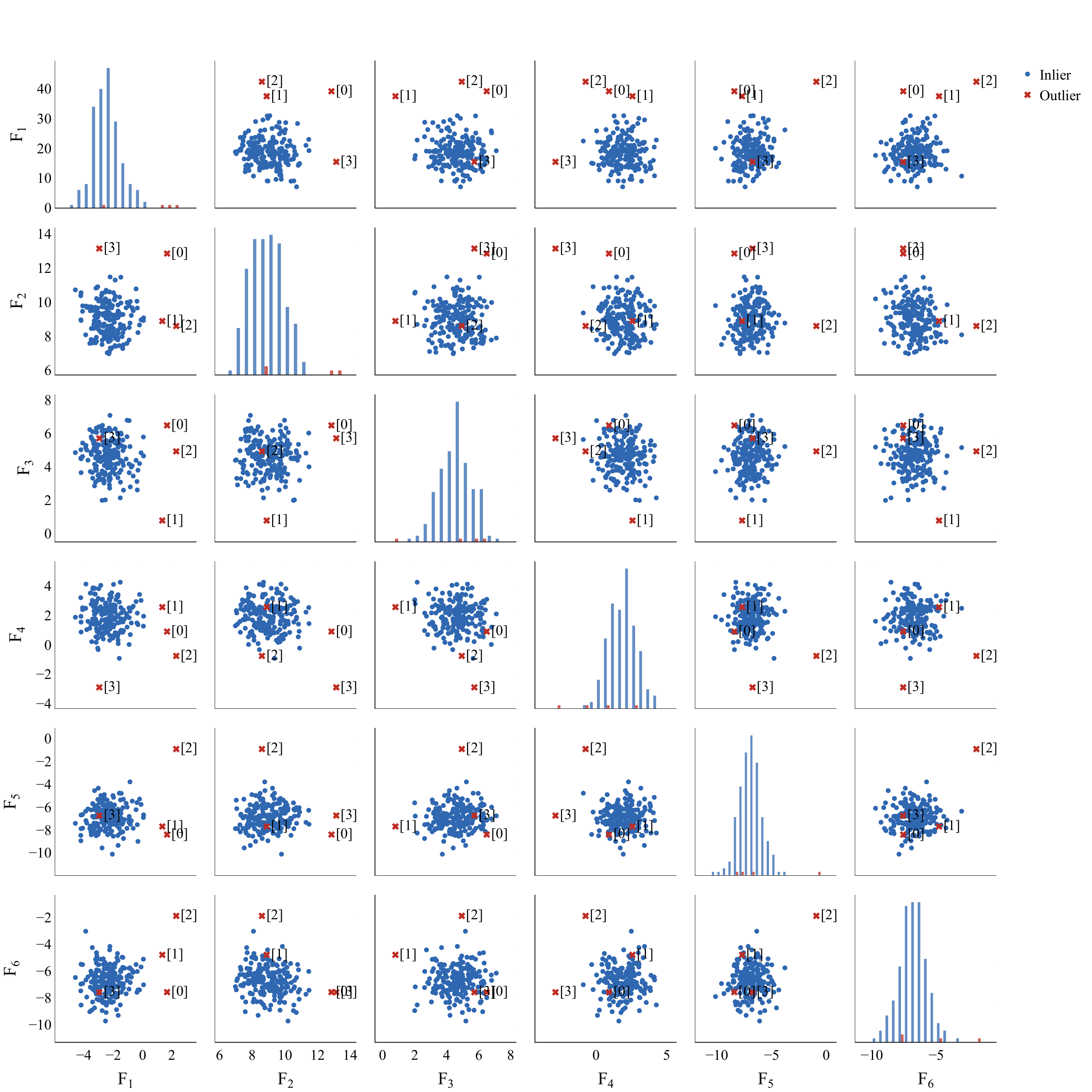}
    \caption{Pairplot of the second synthetic dataset. The dataset comprises \num{200} samples with six numerical features, and four samples have been modified by altering between two to four.}
    \label{fig:pp_d2_overview}
\end{figure}

The second dataset is created by modifying four randomly selected samples according to the previously described rule, as detailed in \Cref{tab:modifiche_d2}.
\Cref{fig:pp_d2_overview} presents an overview of the entire dataset, highlighting four outliers. Unlike the previous dataset, this one is more complex because each outlier is generated by modifying different features. As a result, each outlier can be individually distinguished by a specific set of features, meaning that no single split can separate all outliers from the inliers.\\

\begin{table}[ht!]
\centering
\begin{tabular}{c@{\hspace{0.5cm}}c@{\hspace{0.5cm}}c@{\hspace{0.5cm}}c@{\hspace{0.5cm}}c}
    \toprule
    Outliers & Feature & Initial Value & Final Value & Alteration \\
    \midrule
    \multirow{2}{*}{Sample 0} & $F_0$ & $-1.86$ & $1.67$  & $+3.53$ \\
                            & $F_1$ & $8.92$ & $12.84$ & $+3.93$ \\
    \midrule
    \multirow{2}{*}{Sample 1} & $F_0$ & $-2.19$ & $1.34$ & $+3.53$ \\
                            & $F_2$ & $4.74$ & $0.78$ & $-3.95$ \\
    \midrule
    \multirow{4}{*}{Sample 2} & $F_0$ & $-2.12$ & $2.29$  & $+4.41$ \\
                            & $F_3$ & $4.05$ & $-0.76$ & $-4.81$ \\
                            & $F_5$ & $-7.21$ & $-1.88$ & $+5.33$ \\
                            & $F_4$ & $-6.01$ & $-0.93$ & $+5.08$ \\
    \midrule
    \multirow{2}{*}{Sample 3} & $F_1$ & $9.21$ & $13.13$ & $+3.93$ \\
                            & $F_3$ & $0.95$ & $-2.90$ & $-3.84$ \\
    \bottomrule
\end{tabular}
\caption{The first column lists the outliers in the second synthetic dataset. The table shows the initial and final values of the modified features for these samples, along with the specific modifications applied to introduce the outliers.}
\label{tab:modifiche_d2}
\end{table}

\begin{figure}[ht]
    \centering
    \begin{minipage}[t]{0.9\textwidth}
        \includegraphics[width=\textwidth, keepaspectratio]{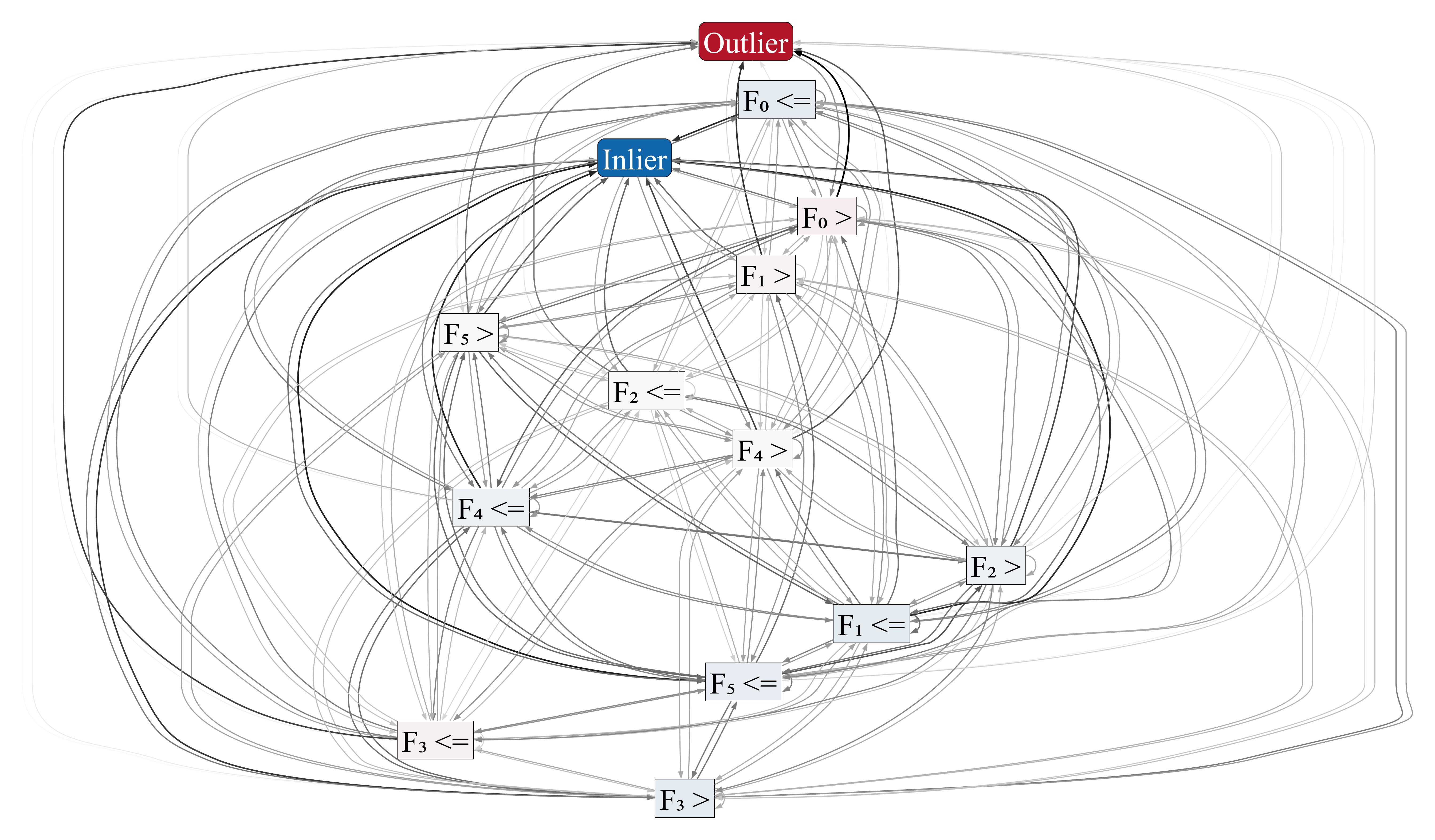}
        \includegraphics[width=\textwidth, keepaspectratio]{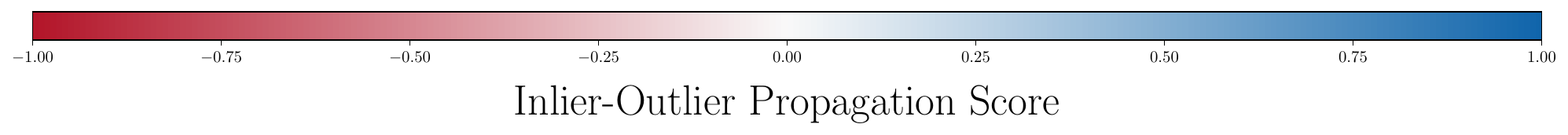}
    \end{minipage}
    \begin{minipage}[t]{0.092\textwidth}
        \raisebox{0cm}{\includegraphics[width=0.9\textwidth, height=8cm, keepaspectratio]{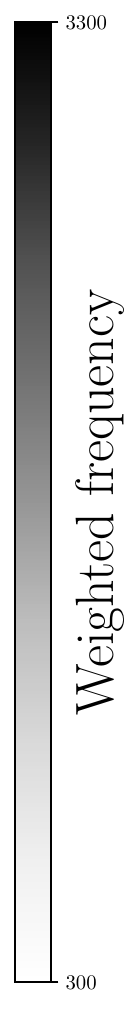}}
    \end{minipage}
    
    \caption{Global representation of the iForest model as a DPG produced by our method for the second synthetic dataset. The vertical bar on the right indicates the edge weights, while the horizontal bar at the bottom displays the IOP-Score of the nodes.}
    \label{fig:global_sintetic_fig_d2}
\end{figure}

The trained iForest model successfully distinguished the modified samples as outliers. 
Moreover, features exhibiting consistent directional changes, such as increases in $F_0$ and $F_1$ or decreases in $F_3$, are more readily distinguishable than others.
Similarly, as for the previous dataset, we applied our technique to explain the iForest process.
The model is converted into the DPG shown in \Cref{fig:global_sintetic_fig_d2}, where the classes ``Outlier'' and ``Inlier'' are distinguished by different colors. For each node, the IOP-Score is computed and represented by its color---these scores are summarized in \Cref{tab:global_sintetic_tab_d2}.

\begin{table}[ht!]
    \centering
    \begin{tabular}{l@{\hspace{0.5cm}}r} 
        \toprule
        Predicate & IOP-Score \\
        \midrule
        $F_1 <=$ & $0.0884$ \\
        $F_3 >$ & $0.0881$ \\
        $F_0 <=$ & $0.0872$ \\
        $F_5 <=$ & $0.0710$ \\
        $F_4 <=$ & $0.0553$ \\
        $F_2 >$ & $0.0542$ \\
        $F_5 >$ & $0.0112$ \\
        $F_2 <=$ & $0.0112$ \\
        $F_4 >$ & $0.0084$ \\
        $F_1 >$ & $-0.0257$ \\
        $F_3 <=$ & $-0.0316$ \\
        $F_0 >$ & $-0.0494$ \\
        \bottomrule
    \end{tabular}
    \caption{IOP-Score assigned to each predicate (node) extracted from the DPG graph of the iForest model for the second synthetic dataset. The scores quantify a node’s propensity to channel data toward the inliers (positive values) or outliers (negative values) class.}
    \label{tab:global_sintetic_tab_d2}
\end{table}

Our technique helps interpret the inner logical process of the iForest model. 
In this scenario, outliers are less distinct from inliers and require the combined influence of multiple features to be isolated, making the model's structure more challenging to interpret than the previous case. 
Nevertheless, our representation and the IOP-Score provide valuable insights.
We can observe that some predicates have an IOP-Score below 0, so they are strongly connected with the ``Outlier'' class. 
In particular, the nodes with the lowest IOP-Score are $F0 >$, $F3 <=$, and $F1 >$, which comprehend the features deliberately altered to create the anomalies; these predicates are critical for the model to distinguish outliers. 
The thicker edges connecting these nodes to the ``Outlier'' class further underscore their frequent use in splits that isolate anomalous data points. 
Moreover, the directional signs in these predicates reveal how the model leverages the features---\Cref{fig:pp_d2_overview} clearly shows that multiple outliers are isolated using these key splits. 
In addition, although $F_4$, $F_2$, and $F_5$ are also modified, their IOP-Scores are slightly above \num{0}, indicating that splits involving these features do not consistently lead to outlier isolation.
Finally, the remaining nodes, with IOP-Score values above \num{0}, are primarily involved in splits that classify points as inliers.

\subsection{Annthyroid dataset}

To evaluate the performance of our XAI methods in a real-world scenario, we used the \emph{Annthyroid dataset}, which is widely adopted in the literature on outlier detection as a benchmark \cite{DVN/OPQMVF_2015,Goldstein2016}.
The dataset represents thyroid function measurements, including hormone levels, biochemical indicators, and patient demographics. Each row corresponds to a patient sample, with multiple attributes capturing relevant physiological parameters. It consists of six numerical features (excluding the binary features) and \num{6916} samples. The features explored include \emph{Age}, which provides demographic context; Thyroid-Stimulating Hormone (\emph{TSH}), a critical regulator of thyroid function; \emph{T3}, \emph{TT4} (Total Thyroxine), and Free Thyroxine Index (\emph{FTI}), which measure hormone concentrations in the blood; Thyroxine Uptake (\emph{T4U}), which helps assess hormone-binding activity. The dataset consists of two classes: \emph{normal} (inliers) and \emph{anomalous} (outliers), where anomalies correspond to thyroid disorders. The class distribution is highly imbalanced, with normal cases forming the majority and anomalous instances accounting for only \SI{3.61}{\percent} of the total samples. The Annthyroid dataset is available in the UCI machine learning repository in the medical domain \cite{bache_uci_nodate}.

\begin{figure}[ht!]
    \centering
    \begin{minipage}[t]{0.9\textwidth}
        \includegraphics[width=\textwidth, keepaspectratio]{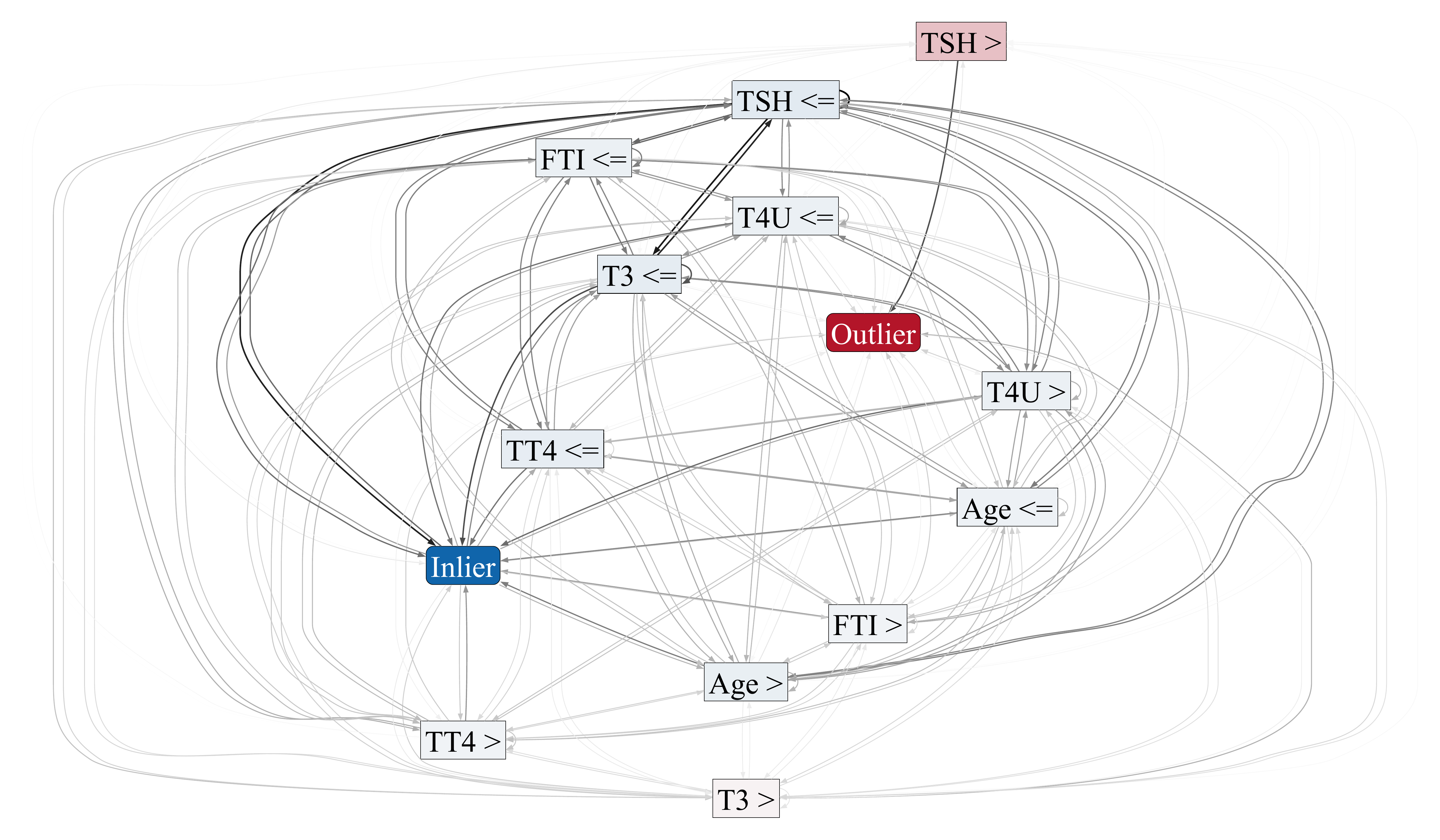}
        \includegraphics[width=\textwidth, keepaspectratio]{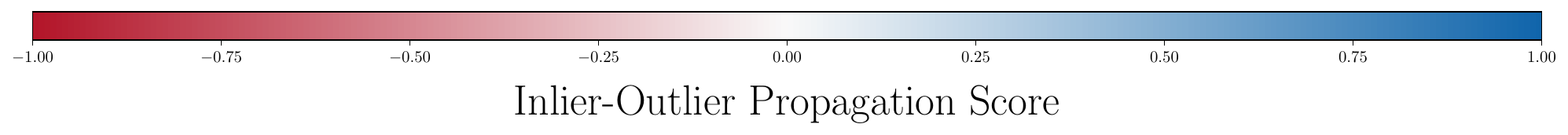}
    \end{minipage}
    \begin{minipage}[t]{0.092\textwidth}
        \raisebox{0cm}{\includegraphics[width=0.9\textwidth, height=8cm, keepaspectratio]{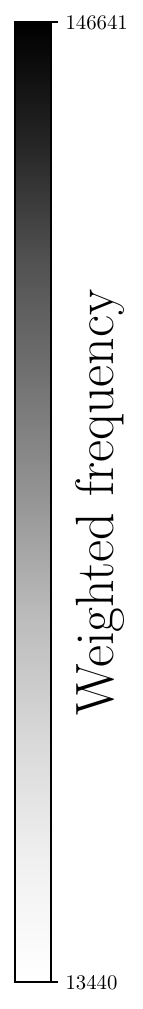}}
    \end{minipage}
    
    \caption{Global representation of the iForest model as a DPG structure produced by our method for the Annthyroid dataset. The vertical bar on the right indicates the edge weights, while the horizontal bar at the bottom displays the IOP-Score of the nodes.}
    \label{fig:global_sintetic_fig_annthyroid}
\end{figure}

We applied our proposal to obtain an iForest model (using \num{200} iTrees) into a DPG and obtained results similar to the literature \cite{Goldstein2016}. The explanation can be appreciated in \Cref{fig:global_sintetic_fig_annthyroid}, where nodes represent predicate-based decision points while edges indicate the flow of decisions through these conditions. Thicker, darker edges correspond to frequently used decision paths, highlighting influential features, whereas lighter edges represent less significant decisions. \emph{TSH} feature serves as a strong predicate point, with a high \emph{TSH} value (\emph{TSH >}) directing the flow toward the outlier node (red box). This indicates that high \emph{TSH} levels are a significant factor in identifying thyroid anomalies with a superior limit. Similarly, a low \emph{TSH} value (\emph{TSH <=}) redirects the flow through additional feature-based decisions before reaching a final classification. The thin edges entering the $\emph{TSH} >$ node also imply that this feature alone is usually sufficient to separate outliers from the rest of the dataset. In contrast, the $\emph{T3} >$ feature necessitates further subdivision.

The IOP-Score, in \Cref{fig:global_sintetic_fig_annthyroid}, represented by the color scale at the bottom, provides further insight into how strongly each predicate affects outlier and inlier identification. Red-shaded paths and nodes indicate a high probability of leading to an outlier classification, while blue-shaded paths and nodes suggest a strong inlier association.
\emph{TSH >} is once again revealed as a highly important factor in anomaly detection. The other predicates make a slight contribution, primarily serving to delineate the boundaries of inlier behavior. More details about the obtained IOP-Score is available in \Cref{tab:global_annthyroid_tab}.

\begin{table}[h!]
    \centering
    \begin{tabular}{l@{\hspace{0.5cm}}r} 
        \toprule
        Predicate & IOP-Score \\
        \midrule
        $\text{TSH} <=$ & $0.0965$ \\
        $\text{T3} <=$ & $0.0846$ \\
        $\text{TT4} <=$ & $0.0776$ \\
        $\text{Age} >$ & $0.0683$ \\
        $\text{T4U} <=$ & $0.0573$ \\
        $\text{FTI} <=$ & $0.0556$ \\
        $\text{T4U} >$ & $0.0551$ \\
        $\text{Age} <=$ & $0.0515$ \\
        $\text{TT4} >$ & $0.0354$ \\
        $\text{FTI} >$ & $0.0323$ \\
        $\text{T3} >$ & $-0.0282$ \\
        $\text{TSH} >$ & $-0.2429$ \\
        \bottomrule
    \end{tabular}
    \caption{IOP-Score assigned to each predicate (node) extracted from the DPG of the iForest model for the Annthyroid dataset. The scores quantify a node’s propensity to channel data toward the ``Inlier'' (positive values) or ``Outlier'' (negative values) class.}
    \label{tab:global_annthyroid_tab}
\end{table}

Notably, as observed in \Cref{tab:global_annthyroid_tab}, the node with the lowest score corresponds to the predicate $\emph{TSH} >$. This is particularly significant, as its highly negative score, along with the thick edge connecting it to the ``Outliers'' class, suggests that the model frequently relies on this feature to isolate anomalies. Similarly, the node containing the predicate $\emph{T3} >$ also has a negative score, though closer to zero, indicating that while it contributes to outlier detection, it often requires additional splits to effectively isolate anomalies. Finally, the remaining nodes, with IOP-Score values above \num{0}, are primarily involved in splits that classify data points as inliers.

\section{Limitations and Extensions}

While the proposed approach comprehensively explains the iForest model using DPG, some limitations must be acknowledged. The transformation of iForest into a graph structure introduces additional computational complexity, mainly when dealing with high-dimensional datasets containing many trees. This complexity also leads to scalability issues, as constructing and analyzing the DPG for large-scale iForest models can be memory-intensive, necessitating optimization techniques for practical deployment. Furthermore, although DPG provides a structured representation of the model, interpreting the graph structure in highly complex datasets requires complementary visualization techniques to enhance clarity. Additionally, while existing XAI methods, such as SHAP and DIFFI provide alternative explanations for iForest, a more in-depth comparison with these techniques is necessary to establish the specific advantages and trade-offs of DPG. Another important consideration is that the method relies on predicates extracted from iForest’s split rules, which may not always capture subtle feature interactions.

To address these challenges, future work will focus on optimizing graph construction techniques, improving scalability, and integrating additional interpretability metrics to enhance the usability of DPG-based explanations. The method aims to identify key features that differentiate outliers from inliers by visualizing decision paths in a graph. The incorporation of predicate signs allows for an interpretation of whether a feature contributes to outlier isolation by surpassing a threshold. Combining the graph structure with the IOP-Score enables a global understanding of the model’s decision-making process, shedding light on important features and their role in detecting anomalies.

\section{Conclusion}

In this work, we introduced a novel approach for explaining the iForest model using DPG. The DPG-based explanation provides a structured and interpretable representation of the outlier detection process. It offers a global perspective on the model’s behavior and logic.
Our approach addresses a gap in the explainability of tree-based ensemble models by extending the capabilities of traditional feature importance methods, such as SHAP and DIFFI, which primarily focus on local or vector-based explanations. The DPG allows for comprehensive visualization of decision paths, enabling users to interpret the isolation logic of iForest with greater clarity. Additionally, introducing the IOP-Score ensures that critical predicates contributing to outlier detection are effectively distinguished from those relevant to inliers.
This paper contributes to the field of XAI by providing a transparent and interpretable method for understanding anomaly detection models, offering a highly extensible approach for accurately identifying outlier behavior.

\section*{Disclosure of Interests}
The authors have no competing interests to declare that are relevant to the content of this article.

\bibliographystyle{unsrtnat}
\bibliography{bibliography}

\end{document}